\documentclass[a4paper]{article}
\usepackage{iwslt18}
\usepackage{times}
\usepackage[utf8x]{inputenc}
\usepackage{latexsym}
\usepackage{multirow}

\usepackage{url}

\usepackage{arydshln}
\title{Using Whole Document Context in Neural Machine Translation}

\name{{\textit{} Valentin Mac\'{e}, Christophe Servan}}
\address{QWANT RESEARCH - 7 Rue Spontini, 75116 Paris, France \\
{\small \tt initial.lastname@qwant.com}
}

\date{}

\begin{document}
\maketitle
\begin{abstract}
    
    In  Machine Translation,  considering  the  document as a whole can help to resolve ambiguities and inconsistencies. In this paper, we propose a simple yet promising approach to add contextual information in  Neural  Machine  Translation. We present a method to add source context that capture the whole document with accurate boundaries, taking every word into account. 
    We provide this additional information to a Transformer model and study the impact of our method on three language pairs.
    The proposed approach obtains promising results in the English-German, English-French and French-English document-level translation tasks. 
    We observe interesting cross-sentential behaviors where the model learns to use document-level information to improve translation coherence. 
\end{abstract}

\section{Introduction}

Neural machine translation (NMT) has grown rapidly in the past years \cite{sutskever2014sequence,vaswani2017attention}. 
It usually takes the form of an encoder-decoder neural network architecture in which source sentences are summarized into a vector representation by the encoder and are then decoded into target sentences by the decoder.
NMT has outperformed conventional statistical machine translation (SMT) by a significant margin over the past years, benefiting from gating and attention techniques.
Various models have been proposed based on different architectures such as RNN \cite{sutskever2014sequence}, CNN \cite{gehring2017convolutional} and Transformer \cite{vaswani2017attention}, the latter having achieved state-of-the-art performances while significantly reducing training time.

However, by considering sentence pairs separately and ignoring broader context, these models suffer from the lack of valuable contextual information, sometimes leading to inconsistency in a translated document.
Adding document-level context helps to improve translation of context-dependent parts. 
Previous study \cite{bawden2017evaluating} showed that such context gives substantial improvement in the handling of discourse phenomena like lexical disambiguation or co-reference resolution.

Most document-level NMT approaches focus on adding contextual information by taking into account a set of sentences surrounding the current pair \cite{tiedemann2017neural, wang2017exploiting,jean2017does,miculicich2018document,zhang2018improving,voita2018context}.
While giving significant improvement over the context-agnostic  versions, none of these studies consider the whole document with well delimited boundaries. The majority of these approaches also rely on structural modification of the NMT model \cite{jean2017does, miculicich2018document, zhang2018improving, voita2018context}.
To the best of our knowledge, there is no existing work considering whole documents without structural modifications.

\textbf{Contribution}: 
We propose a preliminary study of a generic approach allowing any model to benefit from document-level information while translating sentence pairs. 
The core idea is to augment source data by adding document information to each sentence of a source corpus. 
This document information corresponds to the belonging document of a sentence and is computed prior to training, it takes every document word into account.
Our approach focuses on pre-processing and consider whole documents as long as they have defined boundaries. 
We conduct experiments using the Transformer base model \cite{vaswani2017attention}.
For the English-German language pair we use the full WMT 2019 parallel dataset.
For the English-French language pair we use a restricted dataset containing the full TED corpus from MUST-C \cite{di-gangi-etal-2019-must} and sampled sentences from WMT 2019 dataset. We obtain important improvements over the baseline and present evidences that this approach helps to resolve cross-sentence ambiguities.



\begin{table*}[t]
\small
\centering
\begin{tabular}{lll}
\hline
\multicolumn{1}{c}{\textit{SOURCE}}   \textit{} & \multicolumn{1}{c}{\textit{TARGET}}                     \\
\textless{}doc1\textgreater \ Pauli is a theoretical physicist          & Pauli est un physicien théoricien                       \\
\textless{}doc1\textgreater \ He received the Nobel Prize               & Il a reçu le Prix Nobel                                 \\
\textless{}doc2\textgreater \ Bees are found on every continent         & On trouve des abeilles sur tous les continents          \\
\textless{}doc2\textgreater \ They feed on nectar using their tongue & Elles se nourrissent de nectar avec leur langue \\
\textless{}doc2\textgreater \ The smallest bee is the dwarf bee         & La plus petite abeille est l'abeille naine    \\         
\hline
\end{tabular}
\caption{Example of augmented parallel data used to train the \textit{Document} model. The source corpus contains document tags while the target corpus remains unchanged.}
\label{fig:docexample}
\end{table*}

\section{Related Work}
Interest in considering the whole document instead of a set of sentences preceding the current pair lies in the necessity for a human translator to account for broader context in order to keep a coherent translation.
The idea of representing and using documents for a model is interesting, since the model could benefit from information located before or after the current processed sentence.

Previous work on document-level SMT started with cache based approaches, \cite{gong2011cache} suggest a conjunction of dynamic, static and topic-centered cache. 
More recent work tend to focus on strategies to capture context at the encoder level.
Authors of \cite{wang2017exploiting} propose an auxiliary context source with a RNN dedicated to encode contextual information in addition to a warm-start  of encoder and decoder states. They obtain significant gains over the baseline.

A first extension to attention-based neural architectures is proposed by \cite{jean2017does}, they add an encoder devoted to capture the preceding source sentence.
Authors of \cite{miculicich2018document} introduce a hierarchical attention network to model contextual information from previous sentences.
Here the attention allows dynamic access to the context by focusing on different sentences and words.
They show significant improvements over a strong NMT baseline.
More recently, \cite{voita2018context} extend Transformer architecture with an additional encoder to capture context and selectively merge sentence and context representations. They focus on co-reference resolution and obtain improvements in overall performances.

The closest approach to ours is presented by \cite{tiedemann2017neural}, they simply concatenate the previous source sentence to the one being translated. While they do not make any structural modification to the model, their method still does not take the whole document into account.

\section{Approach}
\label{sec:approach}

We propose to use the simplest method to estimate document embeddings.
The approach is called SWEM-aver (Simple Word Embedding Model -- average) \cite{shen-etal-2018-baseline}. 
The embedding of a document $k$ is computed by taking the average of all its $N$ word vectors (see Eq. \ref{equa:doc_emb}) and therefore has the same dimension. Out of vocabulary words are ignored.

 \begin{equation}
     Doc_k=\frac{1}{N}\sum^N_{i=1} w_{i,k}
     \label{equa:doc_emb}
 \end{equation}

Despite being straightforward, our approach raises the need of already computed word vectors to keep consistency between word and document embeddings. 
Otherwise, fine-tuning embeddings as the model is training would shift them in a way that totally wipes off the connection between document and word vectors.

To address this problem, we adopt the following approach: First, we train a baseline Transformer model (noted \textit{Baseline} model) from which we extract word embeddings. 
Then, we estimate document embeddings using the SWEM-aver method and train an enhanced model (noted \textit{Document} model) benefiting from these document embeddings and the extracted word embeddings. 
During training, the \textit{Document} model does not fine-tune its embeddings to preserve the relation between words and document vectors. 
It should be noted that we could directly use word embeddings extracted from another model such as Word2Vec \cite{NIPS2013_5021}, in practice we obtain better results when we get these vectors from a Transformer model.
In our case, we simply extract them from the \textit{Baseline} after it has been trained.

Using domain adaptation ideas \cite{sennrich2016controlling,chu2017empirical,chu2018survey}, 
we associate a tag to each sentence of the source corpus, which represents the document information. 
This tag takes the form of an additional token placed at the first position in the sentence and corresponds to the belonging document of the sentence (see Table \ref{fig:docexample}). The model considers the tag as an additional word and replace it with the corresponding document embedding. The \textit{Baseline} model is trained on a standard corpus that does not contain document tags, while the \textit{Document} model is trained on corpus that contains document tags.

The proposed approach requires strong hypotheses about train and test data. 
The first downfall is the need for well defined document boundaries that allow to mark each sentence with its document tag. 
The second major downfall is the need to compute an embedding vector for each new document fed in the model, adding a preprocessing step before inference time.

\section{Experiments}
\label{sec:XP}

We consider two different models for each language pair: the  \textit{Baseline} and the \textit{Document} model. 
We evaluate them on 3 test sets and report BLEU and TER scores. All experiments are run 8 times with different seeds, we report averaged results and p-values for each experiment. 

Translation tasks are English to German, proposed in the first document-level translation task at WMT 2019 \cite{barrault-EtAl:2019:WMT}, English to French and French to English, following the IWSLT translation task \cite{CettoloIWSLT2015}.

\subsection{Training and test sets}

Table \ref{tab:data1} describes the data used for the English-German language pair.
These corpora correspond to the WMT 2019 document-level translation task. 
Table \ref{tab:data2} describes corpora for the English-French language pair, the same data is used for both translation directions.

\begin{table}[h]
\normalsize
    \centering
    \begin{tabular}{lrrr}
    \hline
Corpora & \#lines & \# EN & \# DE\\
    \hline
Common Crawl & 2.2M & 54M & 50M\\
Europarl V9\textsuperscript{†} & 1.8M & 50M & 48M\\
News Comm. V14\textsuperscript{†} & 338K & 8.2M & 8.3M\\
ParaCrawl V3 & 27.5M & 569M & 527M\\
Rapid 19\textsuperscript{†} & 1.5M & 30M & 29M\\
WikiTitles & 1.3M & 3.2M & 2.8M\\
    \hline
Total Training  & 34.7M & 716M & 667M\\
    \hline
newstest2017\textsuperscript{†} & 3004 & 64K & 60K\\
newstest2018\textsuperscript{†} & 2998 & 67K & 64K\\
newstest2019\textsuperscript{†} & 1997 & 48K & 49K\\
    \hline
    \end{tabular}
    \caption{Detail of training and evaluation sets for the English-German pair, showing the number of lines, words in English (EN) and words in German (DE). Corpora with document boundaries are denoted by †.}
    \label{tab:data1}
\end{table}

\begin{table}[h]
\normalsize
    \centering
    \begin{tabular}{lrrr}
    \hline
Corpora & \#lines & \# EN & \# FR\\
    \hline
News Comm. V14\textsuperscript{†} & 325K & 9.2M & 11.2M\\
ParaCrawl V3 (sampled) & 3.1M & 103M & 91M\\
TED\textsuperscript{†} & 277K & 7M & 7.8M\\
    \hline
Total Training  & 3.7M & 119.2M & 110M\\
    \hline
tst2013\textsuperscript{†} & 1379 & 34K & 40K\\
tst2014\textsuperscript{†} & 1306 & 30K & 35K\\
tst2015\textsuperscript{†} & 1210 & 28K & 31K\\
    \hline
    \end{tabular}
    \caption{Detail of training and evaluation sets for the English-French pair in both directions, showing the number of lines, words in English (EN) and words in French (FR). Corpora with document boundaries are denoted by †.}
    \label{tab:data2}
\end{table}

For the English-German pair, only 10.4\% (3.638M lines) of training data contains document boundaries. For English-French pair, we restricted the total amount of training data in order to keep 16.1\% (602K lines) of document delimited corpora.
To achieve this we randomly sampled 10\% of the ParaCrawl V3. 
It means that only a fraction of the source training data contains document context. The enhanced model learns to use document information only when it is available.

All test sets contain well delimited documents, \textit{Baseline} models are evaluated on standard corpora while \textit{Document} models are evaluated on the same standard corpora that have been augmented with document context.
We evaluate the English-German systems on newstest2017, newstest2018 and newstest2019 where documents consist of newspaper articles to keep consistency with the training data. 
English to French and French to English systems are evaluated over IWSLT TED tst2013, tst2014 and tst2015 where documents are transcriptions of TED conferences (see Table \ref{tab:data2}). 

\begin{table*}[t!]
    \centering
    \begin{tabular}{llllllll}
\hline 
\textbf{Model}  &    & \multicolumn{2}{c}{newstest2017} &  \multicolumn{2}{c}{newstest2018} &  \multicolumn{2}{c}{newstest2019} \\
  En$\rightarrow$De &  & BLEU & TER & BLEU & TER & BLEU & TER\\
\hline 
Baseline &  & 26.78 & 54.82 & 40.61 & 41.02 & 35.67 & 46.80\\
\hline 
Document &  & \textbf{26.96}\textsuperscript{∗∗} & \textbf{54.76} & \textbf{40.77} & \textbf{40.97} & \textbf{36.52}\textsuperscript{∗} & \textbf{46.36}\textsuperscript{∗}\\
\hline
    \end{tabular}
    \caption{Results obtained for the English-German translation task, scored on three test sets using BLEU and TER metrics. p-values are denoted by * and correspond to the following values: \textsuperscript{∗}\textless \ .05,  \textsuperscript{∗∗}\textless \  .01, \textsuperscript{∗∗∗}\textless \ .001.}
    \label{tab:results1}
\end{table*}

\begin{table*}[t!]
    \centering
    \begin{tabular}{llllllllll}
\hline 
Translation  & \multirow{2}{*}{\textbf{Model}}  &  \multicolumn{2}{c}{tst2013} &  \multicolumn{2}{c}{tst2014} &  \multicolumn{2}{c}{tst2015} \\
  direction & &  BLEU  & TER & BLEU & TER & BLEU & TER\\
\hline 
\multirow{2}{*}{En$\rightarrow$Fr} & Baseline & 46.05 & 37.83 & 43.38 & 39.71 & 41.41 & 42.18\\
  & Document &\textbf{46.53}\textsuperscript{∗} & \textbf{37.15}\textsuperscript{∗∗} & \textbf{44.14}\textsuperscript{∗∗} & \textbf{38.95}\textsuperscript{∗∗} & \textbf{42.50}\textsuperscript{∗∗∗} & \textbf{41.33}\textsuperscript{∗∗∗}\\
\hline 
\multirow{3}{*}{Fr$\rightarrow$En} & Baseline & 45.99 & 34.64 & 42.96 & 37.30 & 39.91 & 39.06\\
& Document+tuning & 45.94 & 34.42 & 43.16 & 36.93 & 40.14 & 38.70\\
& Document &  \textbf{47.28}\textsuperscript{∗∗∗} & \textbf{33.80}\textsuperscript{∗∗∗} & \textbf{44.46}\textsuperscript{∗∗∗} & \textbf{36.34}\textsuperscript{∗∗∗} & \textbf{41.72}\textsuperscript{∗∗∗} & \textbf{38.04}\textsuperscript{∗∗∗}\\
\hline
    \end{tabular}
    \caption{Results obtained for the English-French and French-English translation tasks, scored on three test sets using BLEU and TER metrics. p-values are denoted by * and correspond to the following values: \textsuperscript{∗}\textless \ .05,  \textsuperscript{∗∗}\textless \  .01, \textsuperscript{∗∗∗}\textless \ .001.}
    \label{tab:results2}
\end{table*}

Prior to experiments, corpora are tokenized using Moses tokenizer \cite{koehn2007moses}. 
To limit vocabulary size, we adopt the BPE subword unit approach \cite{sennrich2016neural}, through the SentencePiece toolkit \cite{kudo2018sentencepiece}, with 32K rules.

\subsection{Training details}

We use the OpenNMT framework \cite{2017opennmt} in its TensorFlow version to create and train our models. 
All experiments are run on a single NVIDIA V100 GPU. 
Since the proposed approach relies on a preprocessing step and not on structural enhancement of the model, we keep the same Transformer architecture in all experiments.
Our Transformer configuration is similar to the baseline of \cite{vaswani2017attention} except for the size of word and document vectors that we set to $d_{model} = 1024$, these vectors are fixed during training.
We use $N = 6$ as the number of encoder layers, $d_{ff} = 2048$ as the inner-layer dimensionality, $h = 8$ attention heads, $d_k = 64$ as queries and keys dimension and $Pdrop = 0.1$ as dropout probability. All experiments, including baselines, are run over 600k training steps with a batch size of approximately 3000 tokens.

For all language pairs we trained a \textit{Baseline} and a \textit{Document} model. 
The \textit{Baseline} is trained on a standard parallel corpus and is not aware of document embeddings, it is blind to the context and cannot link the sentences of a document.
The \textit{Document} model uses extracted word embeddings from the \textit{Baseline} as initialization for its word vectors and also benefits from document embeddings that are computed from the extracted word embeddings. 
It is trained on the same corpus as the \textit{Baseline} one, but the training corpus is augmented  with (see Table \ref{fig:docexample}) and learns to make use of the document context.

The \textit{Document} model does not consider its embeddings as tunable parameters, we hypothesize that fine-tuning word and document vectors breaks the relation between them, leading to poorer results. 
We provide evidence of this phenomena with an additional system for the French-English language pair, noted \textit{Document+tuning} (see Table \ref{tab:results2}) that is identical to the \textit{Document} model except that it adjusts its embeddings during training.

The evaluated models are obtained by taking the average of their last 6 checkpoints, which were written at 5000 steps intervals. 
All experiments are run 8 times with different seeds to ensure the statistical robustness of our results. 
We provide \textit{p-values} that indicate the probability of observing similar or more extreme results if the \textit{Document} model is actually not superior to the \textit{Baseline}.

\subsection{Results}

\begin{table*}[t]
    \centering
    \begin{tabular}{lll}
\hline
\textbf{Fr-En} \\
\hline
        


\multirow{3}{*}{Context} &   [...] et quand ma fille avait quatre ans, nous avons regardé "Le Magicien d'Oz" ensemble. \\
        &   Ce film a complètement captivé son imagination pendant des mois. \\
        &   Son personnage préféré était Glinda, bien entendu. \\
\hdashline
Source &  Ça lui donnait une bonne excuse pour porter une robe à paillettes et avoir une baguette magique. \\
Ref. & It gave her a great excuse to wear a sparkly dress and carry a wand. \\
\hdashline
Baseline &  It gave \textbf{him} a good excuse to wear a glitter dress and have a magic wand. \\
Document &  It gave \textbf{her} a good excuse to wear a glitter dress and have a magic wand. \\
\hline

\multirow{3}{*}{Context} &   Mon père passait souvent les grandes vacances à essayer de me guérir ... \\
        &   Mais nous avons trouvé un remède miracle : le yoga. \\
        &   [...] j'étais une comique de stand-up qui ne tenait pas debout. \\
\hdashline
Source &  Maintenant, je peux faire le poirier. \\
Ref. &  And now I can stand on my head. \\
\hdashline
Baseline &  Now I can \textbf{do the pear}. \\
Document &  Now, I can \textbf{wring}. \\
\hline

\multirow{3}{*}{Context} &  C'est le but ultime de la physique : décrire le flux de conscience. \\
& Selon cette idée, c'est donc la conscience qui met le feu aux équations. \\
& Selon cette idée, la conscience ne pendouille pas en dehors du monde physique ... \\
\hdashline
Source &  Elle siège bien en son cœur. \\
Ref. &  It's there right at its heart. \\
\hdashline
Baseline &  \textbf{She} sits well in \textbf{her} heart. \\
Document &  \textbf{It} sits well in \textbf{its} heart . \\
\hline


    \end{tabular}
    \caption{Translation examples for the French-English pair. We took the best models of all runs for both the \textit{Baseline} and the \textit{Document} enhanced model}
    \label{tab:example1}
\end{table*}

\begin{table*}[t!]
    \centering
    \begin{tabular}{ll}
\hline
\textbf{En-Fr} \\
\hline
\multirow{3}{*}{Context} &   [The speaker in this example is an old police officer saving a man from suicide] \\
        &   But I asked him, "What was it that made you come back and give hope and life another chance ?" \\
        &   And you know what he told me ? \\
        
\hdashline
Source &    He said "You listened."\\
Ref. & Il a dit : "Vous avez écouté."\\
\hdashline
Baseline & Il a dit : "\textbf{Tu} as écouté." \\
Document & Il a dit : "\textbf{Vous} avez écouté."  \\
\hline

    \end{tabular}
    \caption{Translation example for the English-French pair.}
    \label{tab:example2}
\end{table*}

Table \ref{tab:results1} presents results associated to the experiments for the English to German translation task, models are evaluated on the newstest2017, neswtest2018 and newstest2019 test sets. Table \ref{tab:results2} contains results for both English to French and French to English translation tasks, models are evaluated on the tst2013, tst2014 and tst2015 test sets.

\textbf{En$\rightarrow$De}: 
The \textit{Baseline} model obtained State-of-The-Art BLEU and TER results according to \cite{bojar2017WMT,bojar2018WMT}.
The \textit{Document} system shows best results, up to 0.85 BLEU points over the \textit{Baseline} on the newstest2019 corpus. 
It also surpassed the \textit{Baseline}e by 0.18 points on the newstest2017 with strong statistical significance, and by 0.15 BLEU points on the newstest2018 but this time with no statistical evidence. 
These encouraging results prompted us to extend experiments to another language pair: English-French.

\textbf{En$\rightarrow$Fr}: 
The \textit{Document} system obtained the best results considering all metrics on all test sets with strong statistical evidence. 
It surpassed the \textit{Baseline} by 1.09 BLEU points and 0.85 TER points on tst2015, 0.75 BLEU points and 0.76 TER points on tst2014, and 0.48 BLEU points and 0.68 TER points on tst2013.

\textbf{Fr$\rightarrow$En}: 
Of all experiments, this language pair shows the most important improvements over the \textit{Baseline}. The \textit{Document} model obtained substantial gains with very strong statistical evidence on all test sets. 
It surpassed the \textit{Baseline} model by 1.81 BLEU points and 1.02 TER points on tst2015, 1.50 BLEU points and 0.96 TER points on tst2014, and 1.29 BLEU points and 0.83 TER points on tst2013.

The \textit{Document+tuning} system, which only differs from the fact that it tunes its embeddings, shows little or no improvement over the \textit{Baseline}, leading us to the conclusion that the relation between word and document embeddings described by Eq. \ref{equa:doc_emb} must be preserved for the model to fully benefit from document context.

\subsection{Manual Analysis}

In this analysis we present some of the many cases that suggest the \textit{Document} model can handle ambiguous situations. 
These examples are often isolated sentences where even a human translator could not predict the good translation without looking at the document, making it almost impossible for the \textit{Baseline} model which is blind to the context.
Table \ref{tab:example1} contains an extract of these interesting cases for the French-English language pair. 

Translation from French to English is challenging and often requires to take the context into account. 
The personal pronoun \textit{"lui"} can refer to a person of feminine gender, masculine gender or even an object and can therefore be translated into \textit{"her"}, \textit{"him"} or \textit{"it"}. 
The first example in Table \ref{tab:example1} perfectly illustrate this ambiguity: the context clearly indicates that \textit{"lui"} in the source sentence refers to \textit{"ma fille"}, which is located three sentences above, and should be translated into \textit{"her"}. 
In this case, the \textit{Baseline} model predict the personal pronoun \textit{"him"} while the \textit{Document} model correctly predicts \textit{"her"}. It seems that the \textit{Baseline} model does not benefit from any valuable information in the source sentence.
Some might argue that the source sentence actually contains clues about the correct translation, considering that \textit{"robe à paillettes"} (\textit{"sparkly dress"}) and \textit{"baguette magique"} (\textit{"magic wand"}) probably refer to a little girl, but we will see that the model makes similar choices in more restricted contexts. 
This example is relevant mainly because the actual reference to the subject \textit{"ma fille"} is made long before the source sentence. 

The second example in Table \ref{tab:example1} is interesting because none of our models correctly translate the source sentence. 
However, we observe that the \textit{Baseline} model opts for a literal translation of \textit{"je peux faire le poirier"} (\textit{"I can stand on my head"}) into \textit{"I can do the pear"} while the \textit{Document} model predicts \textit{"I can wring"}. 
Even though these translations are both incorrect, we observe that the \textit{Document} model makes a prediction that somehow relates to the context: a woman talking about her past disability, who has become more flexible thanks to yoga and can now twist her body.

The third case in table \ref{tab:example1} is a perfect example of isolated sentence that cannot be translated correctly with no contextual information. 
This example is tricky because the word \textit{"Elle"} would be translated into \textit{"She"} in most cases if no additional information were provided, but here it refers to \textit{"la conscience"} (\textit{"consciousness"}) from the previous sentence and must be translated into \textit{"It"}. 
As expected the \textit{Baseline} model does not make the correct guess and predicts the personal pronoun \textit{"She"} while the \textit{Document} model correctly predicts \textit{"It"}. 
This example present a second difficult part, the word \textit{"son"} from the source sentence is ambiguous and does not, in itself, inform the translator if it must be translated into \textit{"her"}, \textit{"his"} or \textit{"its"}. 
With contextual information we know that it refers to \textit{"[le] monde physique"} (\textit{"[the] physical world"}) and that the correct choice is the word \textit{"its"}. 
Here the \textit{Baseline} incorrectly predicts \textit{"her"}, possibly because of its earlier choice for \textit{"She"} as the subject. The \textit{Document} model makes again the correct translation. 

According to our results (see Table \ref{tab:results2}), the English-French language pair also benefits from document-level information but to a lesser extent. For this language pair, ambiguities about personal pronouns are less frequent. Other ambiguous phenomena like the formal mode (use of \textit{"vous"} instead of \textit{"tu"}) appear. Table\ref{tab:example2} presents an example of this kind of situation where the word \textit{"You"} from the source sentence does not indicate if the correct translation is \textit{"Vous"} or \textit{"Tu"}. However it refers to the narrator of the story who is an old police officer. In this case, it is very likely that the use of formal mode is the correct translation. The \textit{Baseline} model incorrectly predicts \textit{"Tu"} and the \textit{Document} model predicts \textit{"Vous"}.

\vspace{-0.2cm}
\section{Conclusion}
In this work, we presented a preliminary study of a simple approach for document-level translation.
The method allows to benefit from the whole document context at the sentence level, leading to encouraging results. 
In our experimental setup, we observed improvement of translation outcomes up to 0.85 BLEU points in the English to German translation task and exceeding 1 BLEU point in the English to French and French to English translation tasks.
Looking at the translation outputs, we provided evidence that the approach allows NMT models to disambiguate complex situations where the context is absolutely necessary, even for a human translator.

The next step is to go further by investigating more elaborate document embedding approaches and to bring these experiments to other languages (e.g.: Asian, Arabic, Italian, Spanish, etc.).
To consider a training corpus with a majority of document delimited data is also very promising.




\bibliographystyle{IEEEtran}
\bibliography{wmt2019}
\end{document}